\documentclass[10pt,twocolumn,letterpaper]{article}

\usepackage[pagenumbers]{iccv} 

\definecolor{iccvblue}{rgb}{0.21,0.49,0.74}
\usepackage[pagebackref,breaklinks,colorlinks,allcolors=iccvblue]{hyperref}
\usepackage{graphicx}
\usepackage{booktabs}
\usepackage{multirow}
\usepackage{hyperref}
\usepackage[accsupp]{axessibility}  
\usepackage{adjustbox}

\def\paperID{7740} 
\def\confName{ICCV}
\def\confYear{2025}

\title{Task-Oriented Human Grasp Synthesis via Context- and Task-Aware Diffusers}

\author{%
    An-Lun Liu\textsuperscript{1} \quad Yu-Wei Chao\textsuperscript{2} \quad Yi-Ting Chen\textsuperscript{1}\thanks{ Corresponding Author} \\[1em]
    \textsuperscript{1}National Yang Ming Chiao Tung University \quad \textsuperscript{2}NVIDIA%
}

\begin{document}
\maketitle
\begin{abstract}
In this paper, we study task-oriented human grasp synthesis, a new grasp synthesis task that demands both task and context awareness.
At the core of our method is the task-aware contact maps. Unlike traditional contact maps that only reason about the manipulated object and its relation with the hand, our enhanced maps take into account scene and task information. This comprehensive map is critical for hand-object interaction, enabling accurate grasping poses that align with the task. We propose a two-stage pipeline that first constructs a task-aware contact map informed by the scene and task. In the subsequent stage, we use this contact map to synthesize task-oriented human grasps. 
We introduce a new dataset and a metric for the proposed task to evaluate our approach. 
Our experiments validate the importance of modeling both scene and task, demonstrating significant improvements over existing methods in both grasp quality and task performance. See our project page for more details: https://hcis-lab.github.io/TOHGS/
\end{abstract}
\section{Introduction}
\label{sec:intro}
Hand-object interaction has been a pivotal topic in computer vision research.
%
We have seen diverse efforts, spanning from hand-object pose estimation~\cite{liu2021semi, hampali2020honnotate}, 3D hand-object reconstruction~\cite{ye2023diffusion, chen2022alignsdf, cao2021reconstructing, hasson19_obman}, human grasp synthesis~\cite{jiang2021hand, wang2023dexgraspnet, tendulkar2023flex, liu2023contactgen, li2022contact2grasp, karunratanakul2020grasping, taheri2020grab, corona2020ganhand}, affordance modeling~\cite{liu2023contactgen, li2022contact2grasp, grady2021contactopt, shan2020understanding, ye2023affordance}, hand-object interaction dataset~\cite{chao2021dexycb, brahmbhatt2019contactdb, brahmbhatt2020contactpose, jian2023affordpose, yang2022oakink, taheri2020grab, hampali2020honnotate, corona2020ganhand, hasson19_obman, FirstPersonAction_CVPR2018}, to hand motion trajectory~\cite{liu2022joint, tan2023egodistill}. 
Predicting realistic and viable hand-object interactions opens up new applications in augmented reality~\cite{10.1145/3491102.3517719}, robotics~\cite{wu2023learning, xu2023unidexgrasp, agarwal2023dexterous, zhu2023toward, qin2022dexmv, mandikal2022dexvip, liu2023dexrepnet, wang2023dexgraspnet, wan2023unidexgrasp++, khargonkar2023neuralgrasps, Bao_2023_CVPR}, and human-robot interactions~\cite{chao2022handoversim, wu2024learning, christen2023synh2r}.

\begin{figure}
    \centering
    \includegraphics[width=1\linewidth]{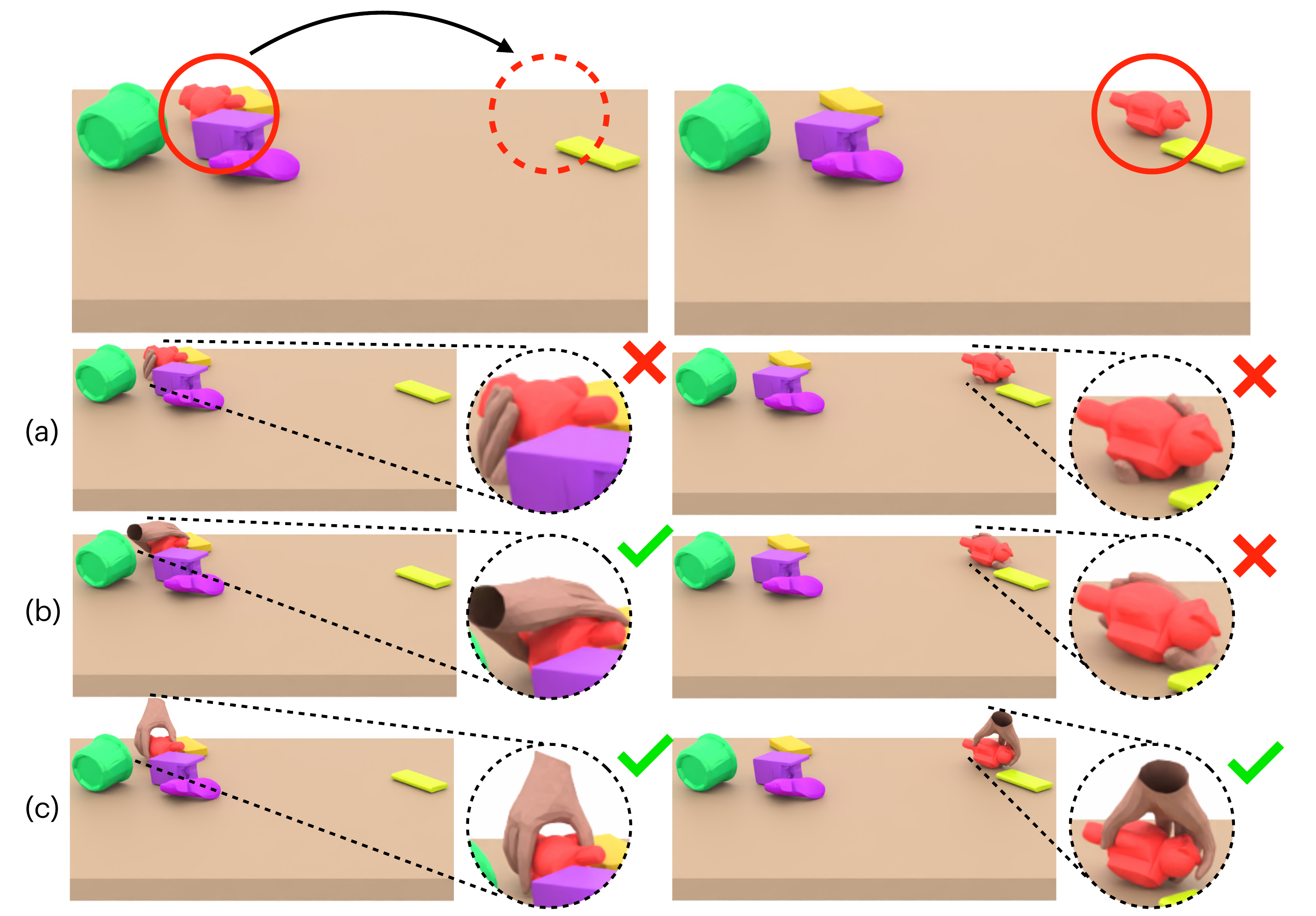}
    \caption{
    Task-oriented human grasp synthesis aims to synthesize collision-free and task-aware human grasps from initial and goal scene point clouds.
    %
    %
    Rows (a) and (b) show failures due to a lack of context or task awareness, causing collisions and penetrations.
    In contrast, (c) demonstrates a successful task-oriented synthesis that considers both scene context and intended task objective.
    %
    }
    \label{fig:f1}
    \vspace{-2mm}
\end{figure}

Despite significant progress in hand-object interaction, limited attention has been given to grasp motion synthesis that both aligns with task objectives and incorporates scene contextual information.
Recent notable work~\cite{christen2022d, goal, CAMS, fullbody_3dv2024, GraspXL, ArtiGrasp} focus on generating plausible human grasp motions for a variety of objects, typically with a single objective, such as moving the object to a desired location~\cite{christen2022d, CAMS, fullbody_3dv2024}. 
GraspXL~\cite{GraspXL} highlights the importance of incorporating multiple motion objectives, such as the desired heading directions and wrist rotations, while ensuring physical plausibility of the motion.
ArtiGrasp~\cite{ArtiGrasp} extends this by considering desired states of objects, such as opening a cabinet door to a target articulation angle or keeping its base steady in its initial pose.
While these approaches show promise in synthesizing grasp motions for specific objectives, they do not consider environmental contexts, such as grasping in a cluttered space for an object relocation task. 
%
This limits their applicability to object-singulated manipulation tasks in less complex environments.

%
To bridge this gap, we study task-oriented human grasp synthesis, a new endeavor that seeks to generate human grasps informed by both scene context and task objectives.
%
%
%
%
%
Fig.~\ref{fig:f1} illustrates a \textit{relocation} task, constituting of grasping a target object and relocating it to a designated location.
%
The task presents two unique challenges.
First, a model needs to possess scene context awareness to ensure that the synthesized grasp avoids collisions with its surroundings.
As shown in the left image of Fig.~\ref{fig:f1} (a), although the synthesized grasp is valid for the object itself, the lack of contextual awareness causes the hand to penetrate the table during the pickup. 
%
Second, the model needs to take the downstream task into account. 
In Fig.~\ref{fig:f1} (b), although the synthesized grasp is successful for the pickup (left image), the lack of task awareness causes the hand to collide with the table during the placing pose (right image).
Finally, Fig.~\ref{fig:f1} (c) illustrates a successful synthesis of a task-oriented human grasp that incorporates both the context and task at hand.

In this work, we first construct a new task-oriented grasp dataset to support the development and evaluation of the problem.
%
%
Despite notable advancements in human grasp synthesis, existing datasets \cite{chao2021dexycb, brahmbhatt2019contactdb, brahmbhatt2020contactpose, jian2023affordpose, yang2022oakink, taheri2020grab, hampali2020honnotate, corona2020ganhand, hasson19_obman, FirstPersonAction_CVPR2018} remain largely focused on object-centric grasp synthesis, overlooking the needs of
downstream hand manipulation tasks, such as stacking objects onto one another.
%
%
We focus on three everyday tasks---\textbf{Placing}, \textbf{Stacking}, and \textbf{Shelving}---all of which require scene awareness to avoid collisions with nearby objects, and task awareness to understand the object's affordance.
For \textbf{Placing} and \textbf{Shelving}, we select 104 everyday objects from DexGraspNet~\cite{wang2023dexgraspnet} including items like bottles, jars, stationery, toys, food, shoes, and 3C electronics. 
Additionally, we create 23 distinct bricks, derived from fundamental geometric shapes, for the \textbf{Stacking} task.
%
%
For each task, we establish a systematic pipeline for generating ground truth human grasps.
%
%
Overall, our dataset contains 571,908 task-oriented human grasps for \textbf{Placing}, 2,989 for \textbf{Stacking}, and 807,028 for \textbf{Shelving}.


Existing human grasp synthesis approaches~\cite{jiang2021hand, liu2023contactgen, grady2021contactopt, li2022contact2grasp} fall short for the proposed task due to two key challenges.
%
%
First, object-centric representations, such as object affordances represented by contact maps~\cite{liu2023contactgen, li2022contact2grasp, grady2021contactopt, jiang2021hand}, fail to account for object-environment interactions, making them prone to collisions in cluttered scenes.
For example, picking and stacking objects in Fig.~\ref{fig:f1} using object-centric representations would result in collisions with surrounding objects, even if the methods produce stable contacts. 
%
Second, current techniques overlook downstream tasks. While the synthesized grasps may be collision-free initially, they can still lead to scene collisions during task execution.
%
%

%

To account for both scene and task contexts, we propose a novel two-stage diffusion-based framework. 
%
%
%
%
%
The framework consists of (1) a \textbf{ContactDiffuser}, which predicts a new representation called task-aware contact map, given the point clouds of the initial and goal scenes along with their corresponding distance maps, and (2) a \textbf{GraspDiffuser}, which synthesizes human grasps based on the predicted task-aware contact map and the object's point cloud.
The rationale for using diffusion models is that task-aware contact maps and their corresponding task-oriented human grasps are inherently multimodal, given the diverse combinations of environmental contexts and tasks.

We conduct extensive experiments on the proposed dataset and demonstrate that our framework, with explicit task-aware contextual awareness, outperforms strong baselines in both physical plausibility and collision avoidance. 
%
%
%
%
Additionally, we empirically show that task-aware contact maps are more effective at extracting task-relevant information than object-centric contact maps.
Our qualitative analysis reveals that the ContactDiffuser generates high-quality contact maps, while the GraspDiffuser produces natural and realistic human grasps.
Furthermore, we perform comprehensive ablation studies to validate the significance of the two proposed diffusers.
Although our work does not involve motion synthesis directly, it represents a crucial step in incorporating task-aware scene awareness into human grasp synthesis. 
Moreover, the generated initial and target human grasps can serve as hand pose references to improve motion synthesis, as demonstrated in~\cite{christen2022d, ArtiGrasp}.
%

Our contributions are summarized as follows. \textbf{First}, we introduce a new task, task-oriented human grasp synthesis, along with a new dataset for its development and benchmarking. \textbf{Second}, we propose a novel two-stage diffusion-based framework that leverages a new representation, the task-aware contact map, to integrate essential scene and task-specific information. \textbf{Third}, we conduct comprehensive experiments to show the effectiveness of our framework over strong baselines.



%

\section{Related Work}

\noindent\textbf{Object Affordance.}
Object offordance has been extensively studied in the community.
Zhu et al.~\cite{Zhu_2015_CVPR} introduce a task-oriented object representation that includes affordance basis, functional basis, imagined actions, and physical concepts.
The affordance basis and functional basis indicate the area to be grasped by the hands and the part intended to act on a target object, respectively.
%
%
The authors of ContactDB~\cite{brahmbhatt2019contactdb} present a dataset that records contact maps, capturing how humans interact with household objects in grasping and handover scenarios using a thermal camera.
AffordPose~\cite{jian2023affordpose} provides part-level affordance annotations for each object, such as twist, pull, and handle-grasp, to enable fine-grained affordance understanding.
The works above aim to ground task-relevant information onto objects to facilitate the generation of corresponding grasps.
However, it becomes challenging for tasks like placing or stacking, where goal scene situations (e.g., cluttered scenes) influence how an object should be grasped.
Rather than predefining affordances as in previous works, we learn task-aware contact maps that simultaneously account for the scene, task, and goal.\\
%
%
%
%
%
%
%
%
%

\begin{figure}[t!]
    \centering
    \includegraphics[width=1\linewidth]{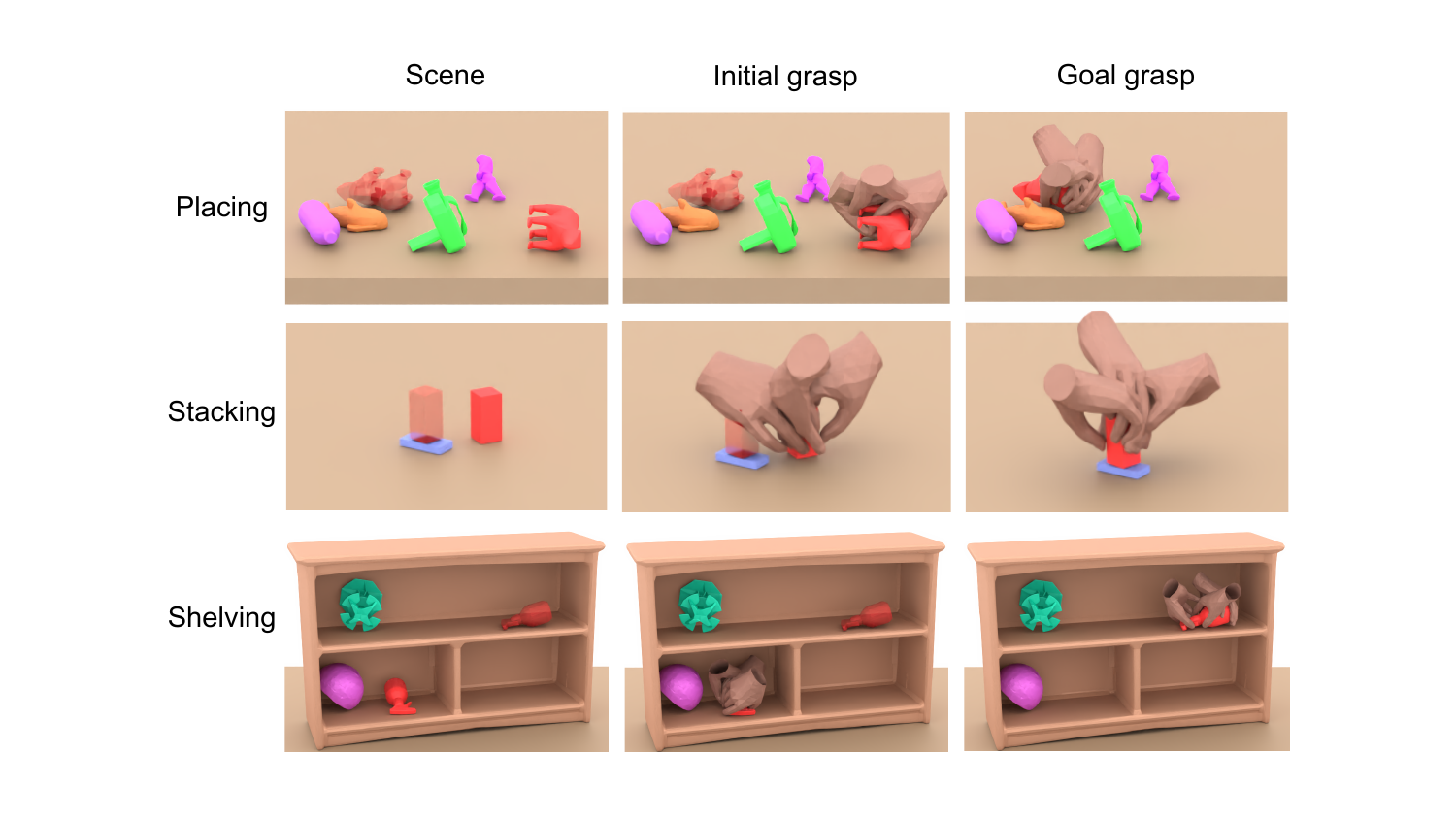}
    \caption{Examples of \textbf{Placing}, \textbf{Stacking}, and \textbf{Shelving}.}
    \label{fig:task}
\end{figure}

\vspace{-1em}
\noindent\textbf{Diffusion Models.}
Diffusion models~\cite{ho2020denoising} have emerged as a powerful generative model and have been leveraged widely across various fields, including generation of images~\cite{saharia2022photorealistic}, video~\cite{videoworldsimulators2024}, human motions~\cite{tevet2023human}, 3D objects~\cite{nichol2022point}, and robot actions~\cite{chi2023diffusion}.
Recently, Huang et al.~\cite{huang2023diffusion} propose SceneDiffuser, a diffusion model-based framework for 3D scene-conditioned human motion, robot motion, and dexterous robotics hand motion generation. 
Inspired by their work, we devise diffusion models for task-oriented human grasp synthesis.

Our work differs from SceneDiffuser in two key aspects. 
First, while SceneDiffuser focuses on generating dexterous robotic hand grasps for object pick-up tasks, our approach targets a range of everyday tasks, such as stacking and shelving.
Second, we explicitly model task-aware scene awareness in human grasp generation, which we demonstrate as essential for producing collision-free grasps that successfully complete tasks.
Additionally, our work complements SceneDiffuser, as our framework can serve as an intermediate understanding step for generating a sequence of hand motions across various tasks in future works.   
%

\label{ch:architecture}
\section{Task-Oriented Human Grasp Generation}

\subsection{Problem Formulation}
Given an initial scene point cloud $S_{\mathtt{init}}$ and a goal scene point cloud $S_{\mathtt{goal}}$, we aim to synthesize task-oriented human grasps ${G}$ that avoid collisions with surrounding objects and successfully accomplish the desired task.
Both scenes contain the target object to be grasped.  
The proposed task is challenging, as it requires modeling both the environmental context and the task simultaneously.

\subsection{Task Specification}

We propose three common tasks to instantiate the proposed problem. 
%
For each task, we generate diverse configurations by simulating the scenes with a physics engine (PyBullet~\cite{coumans2021}).
The three tasks are illustrated in Fig.~\ref{fig:task}.\\
%
\vspace{-1em}

\noindent \textbf{Placing.} 
A target object is placed on a cluttered table and should be picked and moved to a new location on the same table.
We randomize the initial and desired positions of the target. 
First, the object is dropped from a height of 10 cm above the table, and the simulator runs for 5 seconds to allow it to settle into a stable pose.
Next, obstacles from our dataset are randomly selected and positioned near the initial and target locations. If no obstacles collide with the grasped object, the configuration is considered valid.
\\

\vspace{-1em}
\noindent \textbf{Stacking.} Two objects---a base object and an object to be stacked---are placed on a table. 
The goal is to stack the latter on top of the former. 
To create diverse configurations, we randomly select bricks from our dataset, assigning one as the base and the other as the object to be stacked. 
After positioning the stacked object on the base, the simulation runs for 5 seconds.
If both objects remain stable at the end, the configuration is considered valid and is saved.
\\

\vspace{-1em}
\noindent \textbf{Shelving.} A target object is placed on a shelf and should be moved to a different location on the same shelf.
The initial and goal positions are selected randomly. The object is then dropped from 5 cm above its initial position, and the simulation runs for 5 seconds to allow the object to settle.
Finally, obstacles are randomly placed on the shelf. If no obstables collide with the object, the configuration is considered valid.
\\

\vspace{-1em}
\subsection{Dataset}
\noindent \textbf{Human Hand Model.}
We use the MANO hand model~\cite{MANO:SIGGRAPHASIA:2017}, which is a 3D mesh consisting of 778 vertices and 1538 faces.
%
%
The model is controlled by a 10-dimensional parameter $\beta$ for hand shape and a 51-dimensional parameter $\theta$ for joint rotations and root translation.
We use a uniform $\beta$ for all the human grasps. 
%
%
%
%
\\

\vspace{-1em}
\noindent \textbf{Object and Scene.}
We select 104 objects from DexGraspNet~\cite{wang2023dexgraspnet} and rescale them to be graspable by a hand. 
These objects are used in the \textbf{Placing} and \textbf{Shelving} tasks, but many of them are not suitable for \textbf{Stacking}.
%
Consequently, we create 23 distinct brick models with simple geometries specifically designed for \textbf{Stacking}. 
%
We assume all objects are asymmetrical. 
The shelf models used in \textbf{Shelving} are selected from ShapeNet~\cite{chang2015shapenet}.
\\

\begin{figure*}[t!] 
    \centering
    \includegraphics[clip, trim=0cm 3cm 0cm 4cm,width=0.9\linewidth]{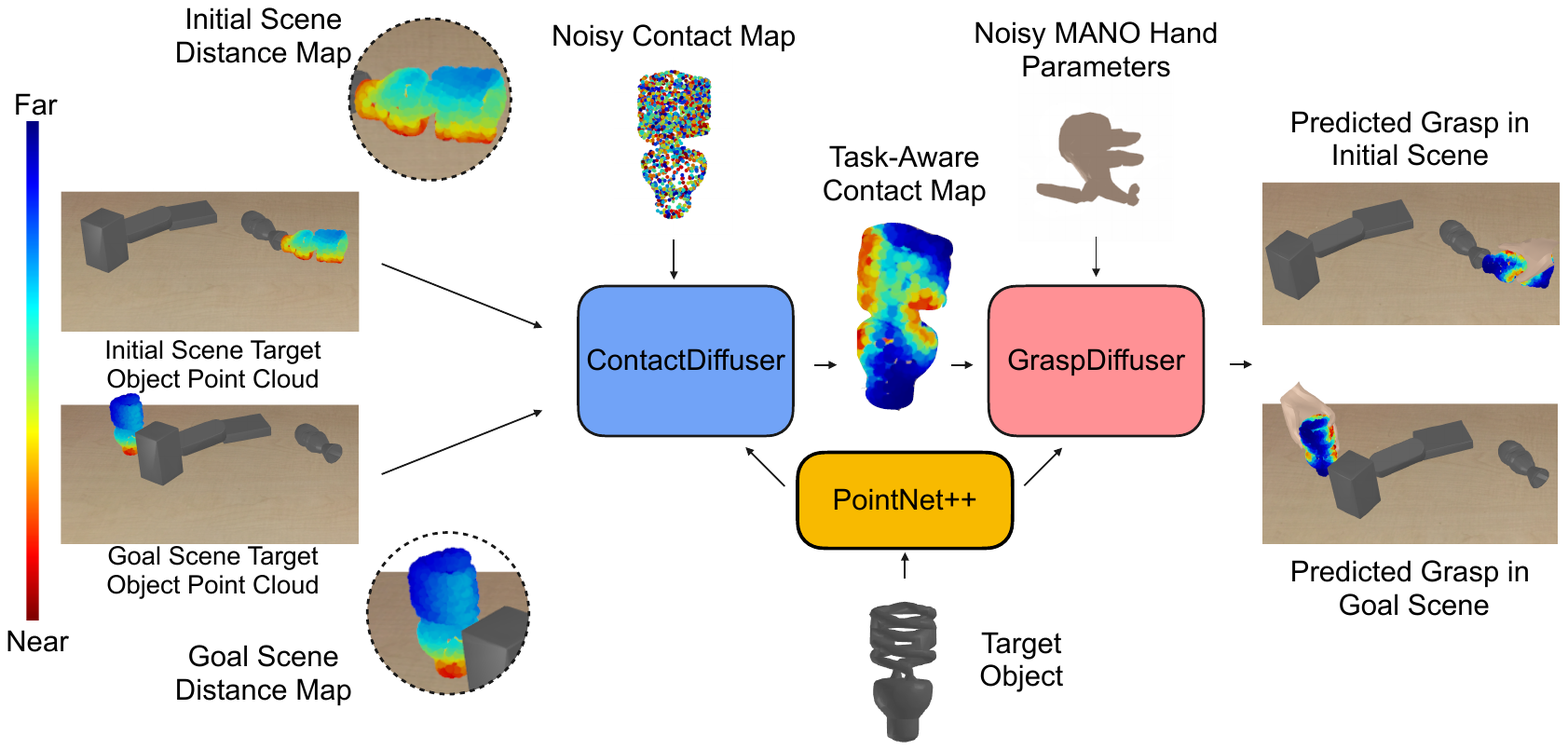}
    \caption{An overview of our proposed task-oriented human grasp synthesis framework. Given 3D initial and goal scenes, our goal is to synthesize task-oriented human grasps that avoid collisions with surrounding objects and achieve the desired task. To capture context- and task-relevant information, our approach integrates the two distance maps, which calculate the shortest distance between the target object and its surroundings in the initial and goal scenes. In addition, we propose to learn a new intermediate representation called task-aware contact maps, integrating environmental context and task-relevant cues to facilitate human grasp synthesis.}
    \label{fig:pipeline}
\end{figure*}

\vspace{-1em}
\noindent \textbf{Dataset Construction.}
We use DexGraspNet~\cite{wang2023dexgraspnet} to generate high-quality human grasps. 
This method leverages a differentiable force closure estimator to efficiently produce diverse and stable grasps at scale. 
However, it may generate grasps that are not physically plausible or human-like. 
To mitigate this, we automatically filter out suboptimal grasps based on penetration volume and simulation displacement---common metrics for assessing grasp quality in grasp synthesis~\cite{karunratanakul2020grasping, jiang2021hand, taheri2020grab, karunratanakul2021skeleton, liu2023contactgen}.
%
%
We set a threshold of \(4 \times 10^{-6} \, \text{cm}^3\) for penetration volume and \(3 \, \text{cm}\) for displacement.
%
%
Subsequently, we manually filter out non-human-like grasps based on joint angles and grasp stability. Finally, we refer to the remaining grasps as prior poses.

%
%

For each target, we sample initial and goal positions, along with their corresponding grasp poses from the prior pose set. Then, we perform collision checks with all the objects in the scene, excluding the object currently grasped by the hand. If the hand remains collision-free throughout the trial, the configuration is considered valid and is saved.
This process expands the prior pose set to over 1.38 million task-oriented grasps.

\section{Methodology}
The overall framework is shown in Fig.~\ref{fig:pipeline}.
Given the initial scene point cloud $S_{\mathtt{init}}$ and the goal scene point cloud $S_{\mathtt{goal}}$, our goal is to synthesize a task-oriented human grasp $G=\{\theta, \beta\}$. The full pipeline consists of two stages. In the first stage, we take the input point clouds and predict an intermediate representation, referred to as the task-aware contact map. In the second stage, the task-aware contact map is used to predict the grasp pose. We treat both stages as generative problems and solve them via two separate diffusion models: \textbf{ContactDiffuser} and \textbf{GraspDiffuser}.

\subsection{Input Preprocessing} 
%
%
Our approach is build upon two types of point cloud based representations: \textit{distance maps} and \textit{task-aware contact maps}. A \textit{distance map} encodes the distance from each point on an object to its closest points in the surrounding scenes. We use two distance maps (i.e., from the initial and goal scene respectively) as the input representation for \textbf{ContactDiffuser}.
Given input point clouds, we randomly sample $N$ points from a target object mesh, where $N$ is set to 2048 following~\cite{huang2023diffusion} to obtain the corresponding target object point cloud $O \in \mathbb{R}^{N \times 3}$. 
To acquire scene point clouds, we merge all points of obstacles (including the table) and perform farthest point sampling to $N_{s}$ points, where $N_{s}$ is 6000. 
We compute the point-to-point shortest distance between a target object point cloud and the scene point clouds. 
We normalize all distance values $d$ to the range of $[0, 1]$. 
To stretch the range of all distance values, we rescale the values via the equation $D = 1 - 2 \times \mathtt{sigmoid}\bigl((\alpha \times d) - 0.5\bigl)$, where the parameter $\alpha$ is set to be 30 in our experiments.
Through the above process, we obtain the two distance maps: the initial scene distance map $D_{\mathtt{init}} \in \mathbb{R}^{N \times 1}$ and the goal scene distance map $D_{\mathtt{goal}} \in \mathbb{R}^{N \times 1}$.
On the other hand, a \textit{task-aware contact map} $\mathcal{C}_{\mathtt{task}} \in \mathbb{R}^{N \times 1}$ encodes the distance from each point on the object to its closest point on the hand mesh model in a grasp pose.
We apply the same process to hand meshes and object point cloud, with an $\alpha$ of 100, to generate ground-truth task-aware contact maps for training and evaluation. 

\subsection{Contact- and Task-aware Diffusers}
%
%

%
%
We use diffusion models~\cite{ho2020denoising} for predicting task-aware contact maps and task-oriented human grasps.
While previous work has applied diffusion models to learn multi-modal distributions for other domains, we adapt them for grasp synthesis and specifically investigate their effectiveness in predicting our proposed representations.

\noindent\textbf{ContactDiffuser.} To capture context- and task-relevant information, our approach leverages the two distance maps, i.e., $D_{\mathtt{init}}$ and $D_{\mathtt{goal}}$, as valuable guidance for generating task-aware contact maps. 
%
%
Task-aware contact maps integrate environmental context (e.g., multiple objects) and task-relevant details (e.g., relocating one object onto another) to facilitate human grasp synthesis. 
The representation contrasts with existing object-centric maps~\cite{li2022contact2grasp, liu2023contactgen, grady2021contactopt, jiang2021hand}, which primarily focus on predicting suitable contact points on an object’s surface for grasping.
%
%

%


We adapt the architecture, a Transformer-based model, proposed in~\cite{huang2023diffusion} for task-aware contact map prediction.
%
%
This architecture processes the current noisy contact map $x_{t}$, timestep $t$, and a set of conditioning variables $C$. 
The conditioning $C$ includes the target object's point cloud and the initial and goal distance maps.
We concatenate each point's initial and goal distances and 3-dimensional coordinates to form a 5-dimensional point cloud.
We then utilize PointNet++~\cite{qi2017pointnet++} to encode the point cloud and extract the corresponding local features. 
%
%
However, the point-order invariance of PointNet++ presents a challenge in mapping features between the point cloud and the contact map.
In natural language processing, sinusoidal positional encoding is often added to retain the sequence order of elements.
%
Instead, we incorporate extracted features as positional embeddings, ensuring that the current noisy contact map $x_{t}$ can be guided to generate the corresponding task-aware contact maps in the denoising process.
%
\\
\noindent\textbf{GraspDiffuser.}
GraspDiffuser takes as inputs the noisy MANO parameters~\cite{mandikal2022dexvip},  target object features extracted via PointNet++, and predicted task-oriented contact maps $\hat{\mathcal{C}}_{\mathtt{task}}$.
First, we concatenate $\hat{\mathcal{C}}_{\mathtt{task}}$ with the target object features to form task-aware object features $\mathcal{O}_{\mathtt{task}}$.
We then apply self-attention~\cite{vaswani2017attention} separately to the noisy MANO parameters~\cite{mandikal2022dexvip} and task-aware object features $\mathcal{O}_{task}$, allowing us to capture the spatial relationships among the various joints and object features, respectively.
Subsequently, we apply cross-attention to establish the correspondence between the object features and the MANO parameters, enabling a more nuanced understanding of the interplay between the object’s geometry and the hand model. 
The design leads to a more accurate prediction of human grasp. 
%
The detailed architectures of ContactDiffuser and GraspDiffuser can be found in the \textcolor{magenta}{Supplementary Material}.
\subsection{Loss Function}
We train the two networks independently. 
The ContactDiffuser network leverages the simplified training objective, denoted as $\mathcal{L}(\theta)$, proposed by Ho et al.,~\cite{ho2020denoising}. 
%
%
For the GraspDiffuser network, we follow previous works~\cite{jiang2021hand, li2022contact2grasp, karunratanakul2021skeleton, liu2023contactgen} that incorporate three additional auxiliary losses to synthesize realistic and physically plausible hand grasp synthesis. 
(1) \textbf{Reconstruction Loss}: the first objective is the mesh reconstruction error. Specifically, we 
calculate the error between the ground truth hand mesh and the predicted hand mesh generated using the MANO hand model. 
The loss is denoted as $L_\mathtt{recon} = \lVert \hat{M} - M \rVert ^2$. 
(2) \textbf{Penetration Loss}: We denote object points that are inside a hand mesh as \( O_{\mathtt{in}} \). The penetration loss is defined as $L_{\mathtt{penetr}} = \frac{1}{|O_{\mathtt{in}}|} \sum_{p \in O_{\mathtt{in}}} \min_{i} \lVert p - \hat{V} _i \rVert _2$, which calculate the distance between an object's point and the closest point on hand vertices. 
(3) \textbf{Consistency Loss}: We enforce that the generated grasps align with the regions the task-aware contact map indicates as graspable. 
The consistency loss is defined as $L_{\mathtt{cons}} = \lVert C - C' \rVert^2$, where \( C \) is the ground truth contact map. 
To obtain \( C' \), we first calculate the shortest distance between each vertex of the target object \( O \) and the vertices of the predicted hand mesh \( \hat{M} \), then normalize these distance maps to form \( C' \), with ranges $[0, \, 1]$.
%
%
The overall training loss of the framework
is defined as follows:
$L = w_{1} \cdot \mathcal{L}(\theta) + w_{2} \cdot L_{\mathtt{recon}} + w_{3} \cdot L_{\mathtt{penetr}} + w_{4} \cdot L_{\mathtt{cons}}$, where the parameters $w_{1}$, $w_{2}$, $w_{3}$, and $w_{4}$ are set empirically as 15, 1, 5, and 0.002, respectively, for \textbf{Placing} and \textbf{Shelving}. For \textbf{Stacking}, they are set as 10, 1, 3, and 0.005, respectively.

\newcommand{\CellWithForceBreak}[2][c]{
\begin{tabular}[#1]{@{}c@{}}#2\end{tabular}}
\section{Experiments}
%
%
Our experiments aim to answer the following questions. 
\textbf{Can the proposed method synthesize high-quality task-oriented human grasps?} 
Despite significant progress in human grasp synthesis~\cite{jiang2021hand, tendulkar2023flex, liu2023contactgen, huang2023diffusion}, we aim to highlight the remaining solution gap. 
%
\textbf{Is the task-aware contact map crucial for task-oriented human grasp synthesis?} 
Modeling contextual and task information is challenging, and we aim to determine whether the proposed task-aware contact map can effectively capture both.
\textbf{Do the proposed ContactDiffuser and GraspDiffuser play crucial roles in synthesizing high-quality task-oriented human grasps?} 
%
%

%

\subsection{Baselines}

\noindent \textbf{Human Grasp Synthesis.} We compare the following human grasp synthesis baselines with GraspDiffuser in our experiments.\\
\noindent\textbf{\underline{GraspTTA}~\cite{jiang2021hand}}: 
GraspTTA utilizes CVAE~\cite{sohn2015learning} to generate an initial coarse human grasp and obtains the final grasp through test-time adaptation.
%

\noindent\textbf{\underline{Modified GraspTTA}~\cite{jiang2021hand}}: We modify GraspTTA by augmenting input point clouds with predicted task-oriented contact maps.

\noindent\textbf{\underline{F-GraspTTA}~\cite{jiang2021hand}}: We estimate the transformation between the initial and goal scene point clouds via iterative closest point.
Next, we use GraspTTA to predict human grasps and compute initial and goal Object Penetration Points (OPPs) for each predicted grasp. If a predicted grasp intersects with either scene—indicated by an OPP exceeding a set threshold—we discard that grasp. The thresholds for Init and Goal OPP are set at 0.05.
%
%
%
%
%
%
%
%
%
However, the method requires accurate pose transformation of the object between initial and goal states. 

\noindent\textbf{\underline{FLEX}~\cite{tendulkar2023flex}}: FLEX is initially designed to generate 3D full-body human grasps. 
%
%
We re-purpose the method to generate human grasps.
Note that the penetration losses in both the initial and goal scenes drive the optimization.
%

\noindent\textbf{\underline{ContactGen}~\cite{liu2023contactgen}}: 
We train ContactGen on our dataset to predict object-centric contact representation, including the contact map, hand-part map, and direction map.
Then, we apply their grasp synthesis solver to obtain predicted grasps.
%
%

\noindent\textbf{\underline{SceneDiffuser}~\cite{huang2023diffusion}}: We reimplement SceneDiffuser~\cite{huang2023diffusion} to predict MANO~\cite{MANO:SIGGRAPHASIA:2017} hand parameters. 
We do not perform optimization during inference, as we apply three auxiliary losses to generate physically plausible human grasps.\\

\vspace{-1em}
\noindent\textbf{Contact Map Generation.} \\
\noindent\textbf{\underline{ContactGen}}~\cite{liu2023contactgen}: We choose ContactGen as a baseline and compare it with the proposed ContactDiffser. Specifically, we adapt the sequential Conditional Variational Autoencoder (CVAE) from ContactGen to generate only a contact map, referring to this variant as \textbf{ContactCVAE}.
%

\subsection{Evaluation Metrics}
For \textbf{Placing} and \textbf{Shelving}, we test on 21 unseen objects. 
As for \textbf{Stacking}, we test on 6 unseen bricks. 
For each object, we test on 10 different task configurations.
For every object in each task configuration, we predict 16 grasps for the evaluation. 
The quality of predicted grasps is evaluated based on their physical plausibility, stability, and collision avoidance, following prior works~\cite{jiang2021hand, taheri2020grab, liu2023contactgen, tendulkar2023flex, karunratanakul2020grasping, karunratanakul2021skeleton}. The following introduces the metrics.
%
%


\noindent\textbf{Penetration Volume (PV):} 
We calculate the penetration volume by converting the meshes into 1mm cubes and measuring the overlap between these voxels.\\
\noindent\textbf{Simulation Displacement (SD):} 
We simulate the object and predicted grasps in PyBullet~\cite{coumans2021} 1 second and then calculate the displacement of the object’s center of mass.\\
\noindent\textbf{Contact Ratio (CR):} Contact percentage of predicted grasps with objects.

\noindent\textbf{Qualified Ratio (QR):} The metric jointly considers both penetration volume and simulation displacement. Notably, a higher penetration volume generally results in lower simulation displacement, which is undesirable. 
We set thresholds at \(3 \times 10^{-6} \, \text{cm}^3\) and \(2 \, \text{cm}\) for penetration volume and simulation displacement, respectively. 
We calculate the percentage of predicted grasps that satisfy both criteria.
%

\noindent\textbf{Diversity Score (DS):} We follow FLEX~\cite{tendulkar2023flex} to compute the average $L_2$ pairwise distance to assess the diversity.

\noindent\textbf{Obstacle Penetration Percentage (OPP):} We compute the penetration percentage of human grasp vertices in obstacles for initial and goal scenes~\cite{tendulkar2023flex}.

\noindent\textbf{Task Score (TS):} We propose a metric called \textbf{TS} to evaluate the quality of task-oriented human grasp synthesis. An effective metric should consider physical plausibility, stability, and collision avoidance in both the initial and goal scenes. Thus, we define \textbf{TS} as 
$\textrm{TS} = \textrm{QR} \times  (1-\textrm{Init\;OPP}) \times  (1-\textrm{Goal\;OPP})$.
%

\begin{table*}[t!]
    \centering
\small
    \setlength{\tabcolsep}{0.3mm}
     \caption{\textbf{Task-oriented Human Grasp Synthesis Evaluation.} \textbf{PV:} Penetration Volume, \textbf{SD:} Simulation Displacement, \textbf{CR:} Contact Ratio, \textbf{QR:} Qualified Ratio, \textbf{OPP:} Obstacle Penetration Percentage, and \textbf{TS}: Task Score. 
%
    %
    Note FLEX~\cite{tendulkar2023flex} struggles to synthesize stable grasps in \textbf{Stacking} due to its inability to handle small bricks.
    %
    }
    \begin{tabular}{l|c|ccccccc|c}
     & Method & PV(Avg/Std)$\downarrow$ & SD(Avg/Std)$\downarrow$ & CR(\%)$\uparrow$ & QR(\%)$\uparrow$ & DS $\uparrow$ & \CellWithForceBreak{ Init\\OPP}(\%)$\downarrow$ & \CellWithForceBreak{Goal\\OPP}(\%)$\downarrow$  & TS$\uparrow$ \\ 
    \hline
    \multirow{5}{*}{Placing} & GraspTTA~\cite{jiang2021hand} & 1.83/2.51 & 2.55/2.90 & 98.92 & 59.10 & 75.01 & 21.36 & 17.67 & 0.382 \\
    & F-GraspTTA~\cite{jiang2021hand} & 1.85/2.29 & 2.47/2.85 & 99.78 & 58.46 & 75.55 & 0.54 & 15.41 & 0.491 \\
    & ContactGen~\cite{liu2023contactgen}& 1.44/1.99 & 3.85/6.75 & 93.57 & 47.29 & \textbf{87.01} & \textbf{5.92} & 17.67 & 0.366\\
    & SceneDiffuser\cite{huang2023diffusion} & \textbf{1.39}/2.04 & 3.10/3.49 & 96.51 & 54.58  & 67.79 & 20.88 & 16.17 & 0.362\\
    & FLEX~\cite{tendulkar2023flex} & 2.61/2.86 & 1.62/2.31 & \textbf{99.88} & 58.09  & 32.92 & 6.84 &  \textbf{5.55} & 0.511\\
    & \textbf{Ours} & 2.36/2.55 & \textbf{1.44}/1.97 & 99.34 & \textbf{64.61} & 41.91 & 6.82 & 6.35 & \textbf{0.563}\\
    \midrule
    \multirow{5}{*}{Stacking} & GraspTTA~\cite{jiang2021hand} & 4.31/2.77 & \textbf{0.28}/0.31 & \textbf{100.00} & 34.87 & 0.30 & 26.05 & 8.32 & 0.236 \\
    & F-GraspTTA~\cite{jiang2021hand} & 0/0 & 0/0 & 0 & 0 & 0 & 0 & 0 & 0 \\
    & ContactGen~\cite{liu2023contactgen} & 0.66/0.66 & 1.85/2.94 & 94.89 & 78.35 & 62.53 & 7.80 & 9.44 & 0.638 \\
    & SceneDiffuser\cite{huang2023diffusion} & 0.53/0.81 & 1.62/2.53 & 94.85 & 78.35 & 63.13 & 25.45 & 9.66 & 0.527\\
    & FLEX~\cite{tendulkar2023flex} & \textbf{0}/0 & 10.65 / 0 & 0/0 & 0.00 & \textbf{107.87} & \textbf{0.5} & \textbf{0} & 0 \\
    & \textbf{Ours} & 1.13/1.22 & 0.97/1.59 & 95.79 & \textbf{84.54} & 48.45 & 14.76 & 4.55 & \textbf{0.687} \\
    \midrule
    \multirow{5}{*}{Shelving} & GraspTTA~\cite{jiang2021hand} & 1.84/2.34 & 2.54/2.91 & 99.04 & 58.86 & 74.63 & 15.70 & 13.61 & 0.428 \\
    & F-GraspTTA~\cite{jiang2021hand} & 1.88/2.08 & 2.68/3.12 & 99.50 & 58.08 & 74.84 & 1.00 & 11.09 & 0.511 \\
    & ContactGen~\cite{liu2023contactgen} & \textbf{1.43}/2.12 & 3.90/4.03 & 93.13 & 46.32 & \textbf{89.79} & 6.42 & 13.17 & 0.376 \\
    & SceneDiffuser\cite{huang2023diffusion} & 1.37/2.20 & 3.35/3.68 & 95.91 & 50.86 & 68.18 & 14.60 & 13.77 & 0.376\\
    & FLEX~\cite{tendulkar2023flex} & 2.80/3.00 & \textbf{1.57}/1.96 & \textbf{99.90} & 57.45 & 29.02 & \textbf{4.41} & \textbf{4.46} & 0.524 \\
    & \textbf{Ours} & 2.15/2.84 & 1.63/2.17 & 99.48 & \textbf{66.94} & 52.74 & 8.55 & 10.06 & \textbf{0.550} \\
    \end{tabular}
    \label{table:t1}
    \vspace{-5mm}
\end{table*}

\subsection{Results and Discussions}
\noindent\textbf{Can the proposed method synthesize high-quality task-oriented human grasps?} We report our empirical studies in Table~\ref{table:t1}. 
%
While GraspTTA, ContactGen, and SceneDiffuser have low penetration volumes (PV$\downarrow$), they struggle with stable grasps (SD$\downarrow$) in \textbf{Placing} and \textbf{Shelving} tasks.
In \textbf{Placing} and \textbf{Shelving}, F-GraspTTA~\cite{jiang2021hand} demonstrates strong performance in the Init \textbf{OPP}. 
However, due to inaccuracies in the transformation obtained from ICP, F-GraspTTA still struggles with the Goal \textbf{OPP}. 
In \textbf{Stacking}, fails to generate suitable human grasps for evaluation, as none of the predicted grasps meet the filtering criteria.
FLEX seeks a balance between penetration volume and simulation displacement. 
ContactGen~\cite{liu2023contactgen} exhibits a lower Init \textbf{OPP} compared to Goal \textbf{OPP}. 
This is attributed to its optimization strategy, which initially orients the hand toward the target object, resulting in a favorable initial \textbf{OPP}. However, due to the lack of scene and task modeling, ContactGen produces a higher goal \textbf{OPP}.

%
%
The \textbf{Stacking} task is more challenging due to the smaller size of the objects compared to those in \textbf{Placing} and \textbf{Shelving}.
GraspTTA~\cite{jiang2021hand} cannot synthesize proper grasps (\textbf{QR}$\uparrow$) and experience severe mode collapse (\textbf{DS}$\uparrow$). 
Similarly, FLEX~\cite{tendulkar2023flex} struggles to achieve stable grasps in the \textbf{Stacking} task due to small bricks.
%
The proposed method demonstrates favorable grasp synthesis (\textbf{QR}$\uparrow$) in the challenging task.
In the \textcolor{magenta}{Supplementary Material}, we provide a thorough analysis of different \textbf{QR} thresholds and their impacts on various baselines.
%

    

\begin{table}[t!]
    \centering
    \scriptsize
    \setlength{\tabcolsep}{0.3mm}
    \caption{A comparison between Object-Centric (OC), Task-Aware (TA), and Ground-Truth (GT) contact maps.}
    \begin{tabular}{l|c|cccc|c}
     & Method & Type &QR(\%)$\uparrow$ &  \CellWithForceBreak{Init\\OPP}(\%)$\downarrow$ & \CellWithForceBreak{Goal\\OPP}(\%)$\downarrow$& TS $\uparrow$\\
    \hline
    \multirow{6}{*}{Placing} & \multirow{3}{*}{M-GraspTTA~\cite{jiang2021hand}} & OC &  51.39 &  19.38 & 16.54 & 0.345 \\ 
    &  & TA &  42.08 & 9.61 & 5.64 & 0.358\\
    \cline{3-7}
    &  & GT &  40.00 & 3.91 & 3.99 & \underline{0.369} \\
    \cmidrule{2-7}
    & \multirow{3}{*}{Ours} & OC & 64.46 & 17.95 & 16.92 & 0.439 \\
    &  & TA & 65.29 & 7.27 & 5.79 & \textbf{0.570} \\
    \cline{3-7}
    &  & GT & 68.30 & 1.69 & 1.66 & \underline{0.660} \\
    
    \midrule
    \multirow{6}{*}{Stacking} & \multirow{3}{*}{M-GraspTTA~\cite{jiang2021hand}} & OC & 90.16 & 26.76 & 9.48 & 0.507\\
    &  & TA & 87.52 & 16.05 & 5.54 & \textbf{0.694}\\
    \cline{3-7}
    &  & GT & 93.92 & 6.67 & 2.70 & \underline{0.852} \\
    \cmidrule{2-7}
    & \multirow{3}{*}{Ours} & OC & 82.77 & 25.05 & 9.51 & 0.561\\
    &  & TA & 84.31 & 14.94 & 4.51 & 0.684 \\
    \cline{3-7}
    &  & GT & 83.33 & 3.12 & 1.11 & 0.798 \\
    \midrule
    \multirow{6}{*}{Shelving} & \multirow{3}{*}{M-GraspTTA~\cite{jiang2021hand}} & OC & 50.66 & 13.80 & 12.44 & 0.382\\
    &  & TA & 52.51 & 11.05 & 11.65 & \textbf{0.412}\\
    \cline{3-7}
    &  & GT & 41.91 & 4.62 & 4.49 & \underline{0.381} \\
    \cmidrule{2-7}
    & \multirow{3}{*}{Ours} & OC & 65.10 & 13.66 & 13.68 & 0.485\\
    &  & TA & 67.49 & 8.72 & 10.31 & \textbf{0.552} \\
    \cline{3-7}
    &  & GT & 67.36 & 4.71 & 4.96 & \underline{0.610} \\
    \end{tabular}
    \label{table:cmap}
    \vspace{-1em}
\end{table}

\begin{table}[h] 
    \centering
    \scriptsize
    \setlength{\tabcolsep}{0.3mm}
    \caption{Ablation study of GraspDiffuser. We denote GraspDiffuser as \textbf{GD}, ContactDiffuser as \textbf{CD}, and Modified-GraspTTA as \textbf{M-GraspTTA}. }
    \begin{tabular}{l|c|ccc|c}
     & Method & QR(\%)$\uparrow$ & \CellWithForceBreak{Init\\OPP}(\%)$\downarrow$ & \CellWithForceBreak{Goal\\OPP}(\%)$\downarrow$& TS $\uparrow$ \\
    \hline
    \multirow{2}{*}{\CellWithForceBreak{Placing\\}}
    & CD + {M-GraspTTA~\cite{jiang2021hand}} & 49.13 & 9.70 & 6.23 & 0.416\\
    & CD + GD & 64.61 & 6.82 & 6.35 & \textbf{0.563}\\
    \midrule
    \multirow{2}{*}{\CellWithForceBreak{Stacking}} 
    & CD + {M-GraspTTA~\cite{jiang2021hand}} & \textcolor{magenta}{86.68} & 15.74 & 5.56 & \textbf{0.689}\\
    & CD + GD & \textcolor{magenta}{84.54} & 14.76 & 4.55 & 0.687 \\
    \midrule
    \multirow{2}{*}{\CellWithForceBreak{Shelving\\}} 
    & CD + {M-GraspTTA~\cite{jiang2021hand}} & 53.11 & 9.56 & 9.45 & 0.434\\
    & CD + GD  & 66.94 & 8.55 & 10.06 & \textbf{0.550} 
    \end{tabular}
    \label{table:t4}
\end{table}

Our method demonstrates a favorable balance between penetration volume and simulation displacement, as reflected in \textbf{QR}. 
Most importantly, it achieves the best performance in \textbf{TS} among all tasks, demonstrating the proposed framework can synthesize high-quality task-oriented human grasps in terms of physical plausibility, stability, and collision avoidance in initial and goal scenes.  
%

\vspace{1mm}
\noindent\textbf{Is the task-aware contact map crucial for task-oriented human grasp synthesis?} 
In Table~\ref{table:cmap}, we compare the influence of different contact maps, i.e., object-centric (OC) and task-aware (TA) contact maps, to the quality of human grasp synthesis. 
Incorporating TA into both Modified-GraspTTA~\cite{jiang2021hand} and our framework results in significant performance improvements in the \textbf{TS} metric compared to using OC.
Additionally, we report results using ground-truth contact maps, which serve as an upper bound for the two-stage framework. 
In particular, GT produces low OPP in the initial and goal scenes.
The observed performance gap highlights a promising research direction for the community to address collectively.
A noteworthy observation is that M-GraspTTA outperforms our method in \textbf{Stacking} when TA is incorporated.
This is attributed to the smaller object size in \textbf{Stacking}. Additional analysis, including qualitative insights, is provided in the following sections.

\begin{figure}[!t]
    \centering    \includegraphics[width=1\linewidth]{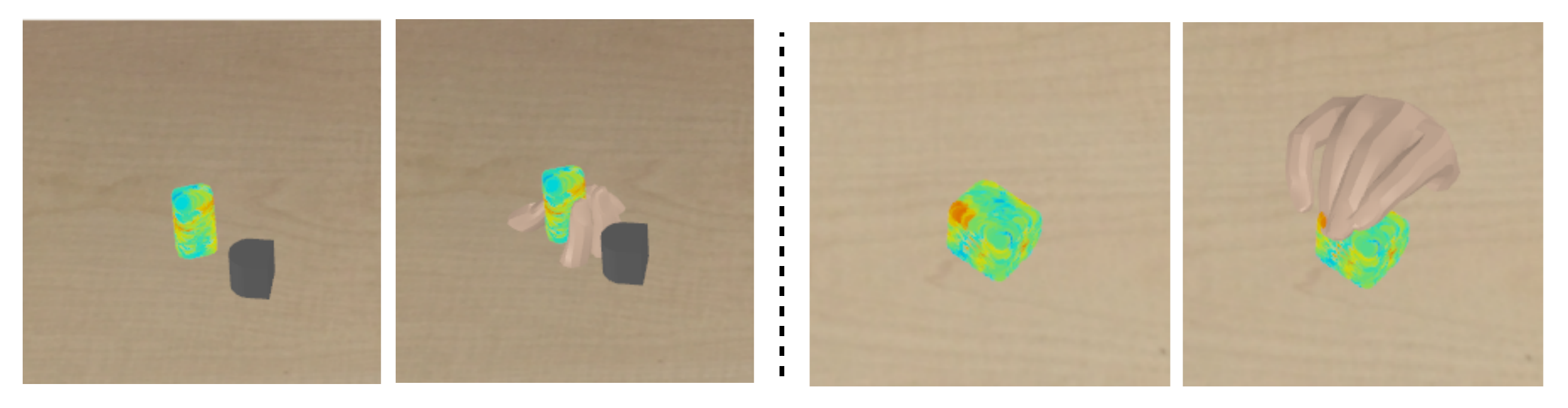}
    \caption{Failure examples. The proposed method struggles with predicting reliable task-aware contact maps for small objects.
}
    \label{fig:failed_grasp}
\end{figure}

\noindent\textbf{Do the proposed ContactDiffuser and GraspDiffuser play crucial roles in synthesizing high-quality task-oriented human grasps?} 
%
Ablation studies of our method are reported in Table~\ref{table:t4} and Table~\ref{table:t5}.
%
%
%
We use ContactDiffuser as the task-aware contact map predictor and study the difference between GraspDiffuser and M-GraspTTA. 
In Table~\ref{table:t4}, we show that GraspDiffuser significantly improves the quality of task-oriented human grasp in \textbf{Placing} and \textbf{Shelving}. 
%
%
In Table~\ref{table:t5}, we use GraspDiffuser as the grasp predictor and study the influence of different contact map predictors.
%
ContactDiffuser generates favorable task-oriented contact maps, resulting in superior task-oriented human grasps in terms of \textbf{TS} for the \textbf{Placing} and \textbf{Shelving} tasks.
%
Note that the proposed method performs on par with other baselines in \textbf{Stacking}, highlighted in \textcolor{magenta}{magenta}. 
This is due to the proposed method cannot produce distinguishable task-oriented contact maps for grasp synthesis, as shown in Fig.~\ref{fig:failed_grasp}.

\subsection{Qualitative Results}
\begin{figure}[!t]
    \centering    \includegraphics[width=0.85\linewidth]{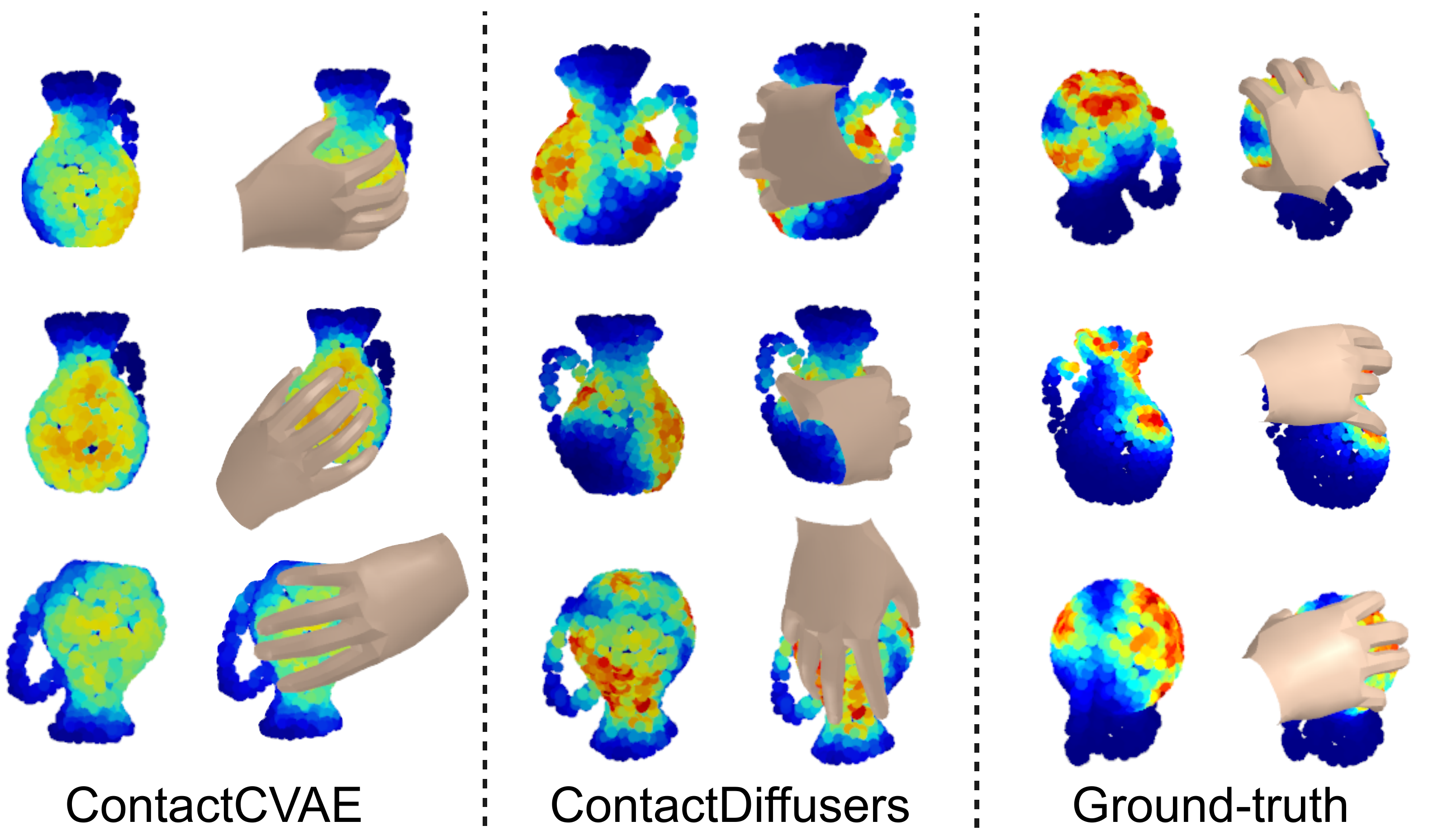}
    \caption{Visualization of predicted contact maps via ContactCVAE~\cite{liu2023contactgen}, ContactDiffuser, the ground-truth contact maps, and the corresponding synthesized human grasps.}
    \label{fig:qualitative_cmap}
\end{figure}

\noindent\textbf{Contact Map.}
Fig.~\ref{fig:qualitative_cmap} shows the results of human grasp synthesis using ContactCVAE~\cite{liu2023contactgen}, ContactDiffuser, and ground truth.
%
ContactCVAE~\cite{liu2023contactgen} tends to predict over-smooth contact maps because it fails to capture multimodal distributions of human grasps under diverse combinations of environmental contexts and tasks.
%
In contrast, ContactDiffuser shows a strong capability to generate contact maps that closely resemble ground-truth contact maps.
%


\begin{table}[t!] 
    \centering
    \scriptsize
    \setlength{\tabcolsep}{0.3mm}
    \caption{Ablation study of ContactDiffuser. We denote GraspDiffuser as \textbf{GD}, ContactDiffuser as \textbf{CD}, and ContactCVAE as \textbf{CC}. }
    \begin{tabular}{l|c|ccc|c}
     & Method & QR(\%)$\uparrow$ & \CellWithForceBreak{Init\\OPP}(\%)$\downarrow$ & \CellWithForceBreak{Goal\\OPP}(\%)$\downarrow$& TS $\uparrow$ \\
    \hline
    \multirow{2}{*}{\CellWithForceBreak{Placing\\}} & CC~\cite{liu2023contactgen} + GD &  62.90 & 7.91 & \textbf{5.14} & 0.549 \\
    & CD + GD &  \textbf{64.61} & 6.82 & 6.35 & \textbf{0.563}\\
    \midrule
    \multirow{2}{*}{\CellWithForceBreak{Stacking}} & CC~\cite{liu2023contactgen} + GD & \textcolor{magenta}{88.68}  & \textbf{15.62} & \textcolor{magenta}{2.89} & \textbf{0.690} \\
    & CD + GD &  \textcolor{magenta}{84.54} & 14.76 & \textcolor{magenta}{4.55} & 0.687 \\
    \midrule
    \multirow{2}{*}{\CellWithForceBreak{Shelving\\}} 
    & CC~\cite{liu2023contactgen} + GD & 59.06 & 11.49 & \textbf{10.19} & 0.469 \\
    & CD + GD &  66.94 & 8.55 & 10.06 & \textbf{0.550} \\
    \end{tabular}
    \label{table:t5}
\end{table}

\begin{figure}[t!]
    \centering    \includegraphics[width=0.95\linewidth]{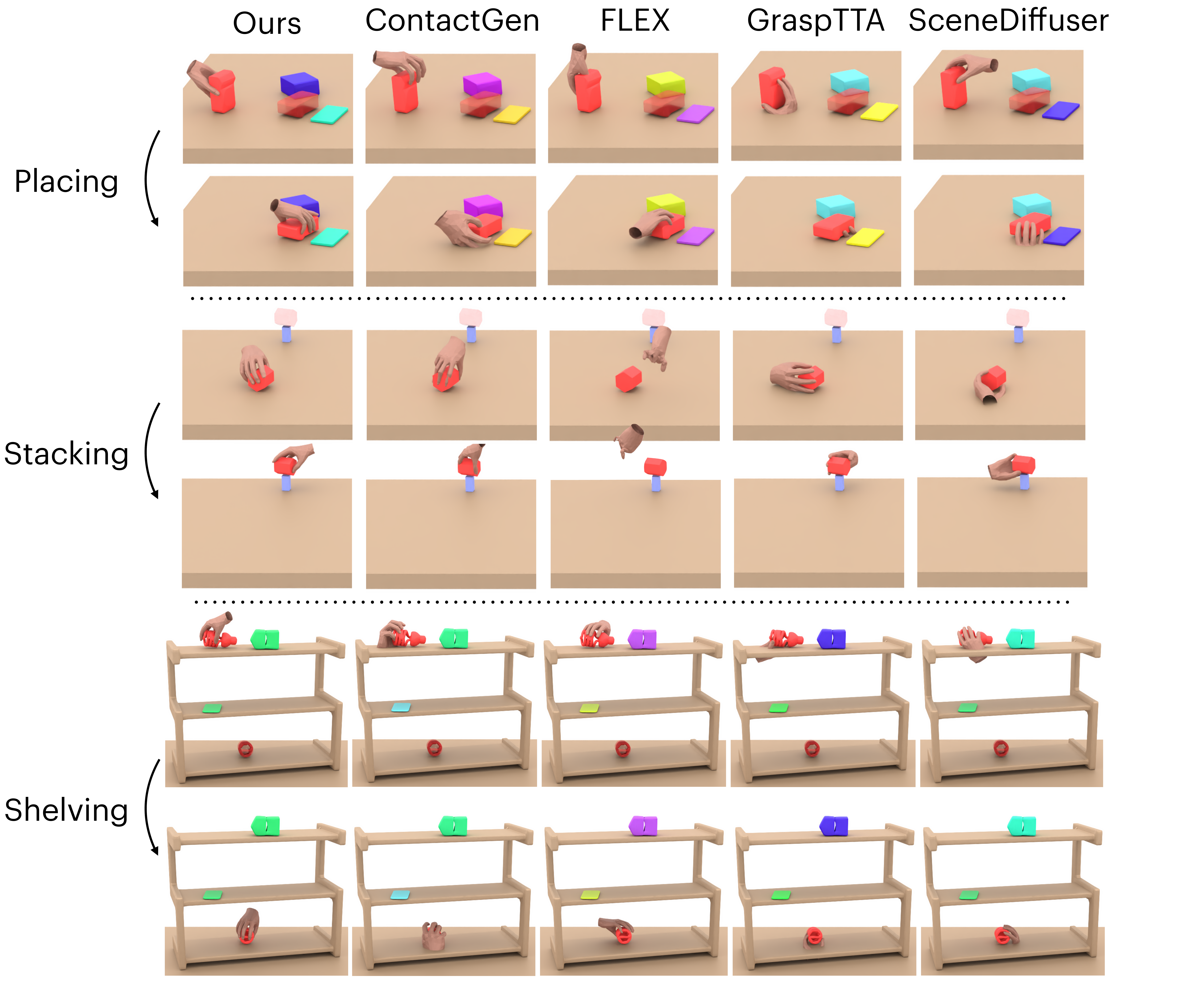}
    \caption{Visualization of predicted human grasps by Ours, ContactGen~\cite{liu2023contactgen}, FLEX~\cite{tendulkar2023flex}, GraspTTA~\cite{jiang2021hand}, and SceneDiffuser~\cite{huang2023diffusion}.}
    \label{fig:qualitative_grasp}
\end{figure}

\noindent\textbf{Synthesized Human Grasps.}
ContactGen~\cite{liu2023contactgen}, GraspTTA\cite{jiang2021hand}, and SceneDiffuser's predictions collide with the scene severely, as shown in Fig.~\ref{fig:qualitative_grasp}. 
FLEX~\cite{tendulkar2023flex} synthesize grasps with unrealistic contact and fail to generate a grasp for \textbf{Stacking}. 
Our method produces high-quality human grasps and avoids collision with obstacles.\\

\vspace{-1em}
\noindent\textbf{Human Study.}  We conduct a human study to evaluate the perceptual quality of different human grasp synthesis methods. 
The study involves 10 subjects without prior research experience in human grasp synthesis.
For \textbf{Placing} and \textbf{Shelving}, we select 3 unseen objects, and for \textbf{Stacking}, we choose 2 unseen bricks. 
For each object, we randomly sampled 8 grasps from the GraspTTA~\cite{jiang2021hand}, FLEX~\cite{tendulkar2023flex}, and Ours.
Subjects are asked to rate each human grasp on a scale from 1 to 6, where 1 indicates grasps that are unnatural, unstable, and collide with the scene, and 6 indicates grasps that are natural, stable, and free of collisions with the scene.
%
Table~\ref{table:human_study} presents each method's mean and standard deviation.
Our method demonstrates promising results on \textbf{Placing} and \textbf{Stacking}, compared to GraspTTA and FLEX. FLEX performs favorably in \textbf{Shelving}. 
This is because human subjects can easily identify penetrations with the scene, resulting in better subjective scores. This observation aligns with the lower OPPs for FLEX in Table~\ref{table:t1}.
%

 \begin{table}[!t]
 \centering
 \scriptsize
    \setlength{\tabcolsep}{0.5mm}
    \caption{Human Study of Synthesized Human Grasps.}
    \begin{tabular}{l|cc|cc|cc}
      &\multicolumn{2}{c}{Placing} &\multicolumn{2}{c}{Stacking} 
      &\multicolumn{2}{c}{Shelving}\\
     Method & Mean$\uparrow$ & Std  & Mean$\uparrow$ & Std & Mean$\uparrow$ & Std\\
    \hline
    GraspTTA~\cite{jiang2021hand} & 2.06 & 0.69 & 2.65 & 0.84 & 2.54 & 0.56\\
    FLEX~\cite{tendulkar2023flex} & 2.65 & 1.03 & 1 & 0 & \textbf{3.73} & 0.82 \\
    Ours & \textbf{3.41} & 0.94 & \textbf{4.43} & 0.93 & 3.38 & 0.55\\
    \midrule
    \end{tabular}
    
    \label{table:human_study}
   
\end{table}
\section{Conculsion}
 In this work, we present a task-oriented human grasp synthesis task and a new dataset for development and benchmarking. 
We show that existing algorithms fail to generate high-quality grasps due to limited context and task modeling. We propose a novel two-stage diffusion-based framework that explicitly integrates essential context and task information to address this.
Our thorough quantitative and qualitative experiments validate our proposed framework's effectiveness compared to strong baselines. 
\\
\noindent\textbf{Limitations and Future Work.}
Our method cannot reliably predict contact maps for small objects, which may lead to collisions. We assume the initial and goal grasps are coherent, though we plan to further evaluate this assumption. At present, our experiments are conducted on a synthetic dataset that provides the goal scene point cloud as input. However, obtaining the target scene point cloud in real-world scenarios may be non-trivial.
We plan to extend our work to the real world and aim to integrate task-oriented human grasp synthesis with hand-object motion synthesis.
%


\noindent\textbf{Acknowledgements.} The work is sponsored in part by the National Science and Technology Council under grants 113-2634-F-002-007-, 113-2628-E-A49-022- and 114-2628-E-A49-007-, the Higher Education Sprout Project of National Yang Ming Chiao Tung University, and the Ministry of Education, the Yushan Fellow Program Administrative Support Grant.

{
    \small
    \bibliographystyle{ieeenat_fullname}
    \bibliography{main}
}

\end{document}